\documentclass[letterpaper, 10 pt, conference]{ieeeconf}  
\IEEEoverridecommandlockouts                              

\usepackage{amsmath,amsfonts}
\usepackage{algorithmic}
\usepackage{algorithm}
\usepackage{array}
\usepackage[caption=false,font=normalsize,labelfont=sf,textfont=sf]{subfig}
\usepackage{textcomp}
\usepackage{stfloats}
\usepackage{url}
\usepackage{verbatim}
\usepackage{graphicx}
\usepackage{hyperref}
\usepackage{booktabs}
\usepackage{multirow}
\hypersetup{hypertex=true,
    colorlinks=true,
    citecolor=green
}
\usepackage{cite}
\usepackage[table]{xcolor}
\usepackage{eso-pic} 

\title{\LARGE \bf
RAM-NAS: Resource-aware Multiobjective Neural Architecture Search Method for Robot Vision Tasks
}

\AddToShipoutPictureFG*{%
  \AtTextUpperLeft{%
    \raisebox{-9.9em}[0pt][0pt]{%
      \parbox{\textwidth}{\centering\footnotesize\itshape
        * Joint first authorship;\ \textdagger\ Corresponding author.}%
    }%
  }%
}

\author{Shouren Mao\textsuperscript{*}, Minghao Qin\textsuperscript{*}, Wei Dong, Huajian Liu, Yongzhuo Gao$^\dagger$ 
\thanks{All authors are with the State Key Laboratory of Robotics and System, Harbin Institute of Technology, Harbin, China. }}

\begin{document}
\newcommand{\TopNotice}{%
  \AddToShipoutPictureFG*{%
    \AtPageUpperLeft{%
      \hspace*{\dimexpr 1in + \hoffset + \oddsidemargin\relax}%
      \raisebox{-2.8em}[0pt][0pt]{%
        \begin{minipage}{\textwidth}
          \centering \fontsize{5.5}{8.6}\selectfont
          © 2024 IEEE. Personal use of this material is permitted. Permission from IEEE must be obtained for all other uses, in any current or future media, including reprinting/republishing this \\ material for advertising or promotional purposes, creating new collective works, for resale or redistribution to servers or lists, or reuse of any copyrighted component of this work in other works.\\[0.5ex]
          {\raggedright\fontsize{8.5}{9.6}\selectfont\bfseries
          \MakeUppercase{Published in IEEE/RSJ IROS 2024 (Abu Dhabi, UAE). DOI: 10.1109/IROS58592.2024.10802271.}\par}
        \end{minipage}%
      }%
    }%
  }%
}
\TopNotice  

\maketitle
\thispagestyle{empty}
\pagestyle{empty}

\begin{abstract}

Neural architecture search (NAS) has shown great promise in automatically designing lightweight models. However, conventional approaches are insufficient in training the supernet and pay little attention to actual robot hardware resources. To meet such challenges, we propose RAM-NAS, a resource-aware multi-objective NAS method that focuses on improving the supernet pretrain and resource-awareness on robot hardware devices. We introduce the concept of subnets mutual distillation, which refers to mutually distilling all subnets sampled by the sandwich rule. Additionally, we utilize the Decoupled Knowledge Distillation (DKD) loss to enhance logits distillation performance. To expedite the search process with consideration for hardware resources, we used data from three types of robotic edge hardware to train Latency Surrogate predictors. These predictors facilitated the estimation of hardware inference latency during the search phase, enabling a unified multi-objective evolutionary search to balance model accuracy and latency trade-offs. Our discovered model family, RAM-NAS models, can achieve top-1 accuracy ranging from $76.7\%$ to $81.4\%$ on ImageNet. In addition, the resource-aware multi-objective NAS we employ significantly reduces the model's inference latency on edge hardware for robots. We conducted experiments on downstream tasks to verify the scalability of our methods. The inference time for detection and segmentation is reduced on all three hardware types compared to MobileNetv3-based methods. Our work fills the gap in NAS for robot hardware resource-aware.

\end{abstract}

\section{INTRODUCTION}

The application of deep learning techniques in robot environmental perception is a current focal point of research. However, their demanding resource requirement in terms of processing time challenges their sustained operation on edge devices embedded within robots\cite{edgeml}. Nevertheless, specific visual perception tasks are driven by deep learning, along with related tasks like object detection\cite{object_detection} and semantic SLAM\cite{vdoslam}.  Consequently, developing lightweight deep learning models becomes crucial for deploying them on edge computing devices embedded within robots. Model lightweighting strives to reduce model parameters while maintaining performance without significant degradation, thus enhancing computational efficiency. Research\cite{han2015learning} suggests that numerous parameters within existing deep neural network models are unnecessary, indicating the potential for model lightweighting. This is particularly crucial for deploying deep learning models on robot platforms. NAS by limiting model complexity, uses reinforcement learning\cite{zoph2016neural} or evolutionary algorithms\cite{evolution_2} to automatically pick and adjust key parameters like network layers, width, and depth based on task performance, aiming for the lightweighting model design. NAS stands out for its automation and innovation, reducing the need for manual design efforts by automatically finding the optimal neural network structure. 

Recently, one-shot NAS \cite{spos} became popular due to low computation overhead and competitive performance. It does not require training thousands of individual models from scratch but only trains a single supernet that contains any architecture in the search space, weights are shared among subnets respectively. However, when a supernet shares weights across subnets, it might not guarantee adequate training for each specific subnet. This inconsistency in ranking the architecture could diminish the efficacy of the search process\cite{bignas}. Retrain-free NAS methods\cite{bignas, attentivenas} propose simultaneously training multiple networks, slicing different-sized subnets directly from the supernet and synchronously training them during the supernet’s training process. 
In this paper, we present subnets mutual distillation and use DKD loss to enable the knowledge better transfer between subnets, without requiring an external teacher model. Through these strategies, the consistency in ranking individual subnets.

On the other hand, previous NAS\cite{fbnet,OFA} mainly abstracts the search problem into a single objective optimization problem under given resource constraints, it demands multiple executions, consuming considerable computing resources and time in the pursuit of a suitable model. Compared to single objective optimization, multiobjective optimization not only allows for simultaneous optimization of multiple given objectives but also yields a Pareto-optimal frontier containing multiple optimal solutions\cite{nsganetv2}. These solutions exhibit different performances across multiple optimization objectives, thereby facilitating more convenient selection and deployment of different robot computational hardware configurations. 
Additionally, these methods\cite{fbnetv2,fbnetv3}, predominantly consider the computing resources of mobile phones, and few take into account the more frequently used edge hardware of robots. However, as the NAS operating on the server side lacks direct latency aware of hardware, inspired by\cite{nsganetv2,tai}, we propose surrogate predictors to assess the architecture's latency. In this paper, we conduct a neural architecture search to be aware of the resources of actual robot edge hardware devices with surrogate-assisted.

In summary, to address the limitation of existing NAS methods, which often overlook robot edge device resources and necessitate multiple executions to identify models with varied computing power requirements, we introduce RAM-NAS.
The key contributions are summarized below:
\begin{itemize}
\item We propose RAM-NAS, a Resource-aware Multiobjective Neural Architecture Search method. This method allows for a single search to compute the Pareto frontier under various constraints, decision makers can deploy models under different resource constraints.
\item We adopt a method where multiple subnets mutually distill each other, enabling joint parameter updates across these subnets. It uses DKD Loss to compute the loss function, ensuring more effective convergence of supernet and boosting the consistency in ranking individual subnets.
\item We adopt reusable surrogate predictors to predict hardware latency,
the predictor demonstrates high sample efficiency and exhibits strong consistency in predicting latency with actual performance.
\end{itemize}

\section{RELATED WORK}
\subsection{Neural Architecture Search}
Early NAS approaches were highly time-consuming. Recent work\cite{spos} has employed the One-Shot Weight Sharing strategy to reduce search costs. Its core idea involves training a supernet where all subnets share weights. However, these methods always require an additional retraining step after determining the optimal architecture. To address this issue, OFA\cite{OFA} introduced a progressive shrinking algorithm that allows full-sized pre-training of the supernet before subnet sampling. However, optimizing all subnets simultaneously during the training of the supernet is not feasible. OFA samples only about $10-12\%$ of the supernet during training, which is not scalable. Inspired by the concept of slim networks\cite{slimmable}, some retrain-free NAS methods propose a novel solution of simultaneously training multiple networks, slicing different-sized subnets directly from the supernet and synchronously training them during the supernet's training process. For instance, BigNAS\cite{bignas} and AttentiveNAS\cite{attentivenas} adopt a training strategy called the "sandwich rule" during supernet training. This strategy selects the largest subnet, the smallest subnet, and N-2 random subnets to perform forward-backward operations and aggregating gradients before updating model weights. However, it only distills other subnets using the largest subnet, failing to utilize the dark knowledge between subnets fully. This paper introduces the subnet Mutual Distillation strategy, enabling mutual distillation among subnets during the super network training phase and utilizing DKD Loss to compute their loss.
\subsection{Resource Aware in NAS}
Recent works have focused more on resource-aware neural architecture search aimed at maximizing predictive accuracy while minimizing resource demands such as latency, FLOPs, or memory usage. For instance, several algorithms\cite{flops_1} incorporate hardware-agnostic computational costs like FLOPs into the loss function to penalize the network's resource consumption on target devices. However, FLOPs don't always reflect actual latency\cite{flops_2}.  Therefore, subsequent works use the actual hardware latency of models as the optimization target. There are two ways to obtain actual hardware latency: one involves acquiring real-time running metrics of the target hardware platform during the search process to guide the search in an online manner, while the other relies on offline perception methods such as proxy models, lookup tables, and more. Online resource-aware method, such as\cite{mnasnet}, measures inference latency on target hardware by executing models on a mobile device. It integrates this measurement into the optimization objective during the search. However, it requires waiting for feedback on resource metrics during the search process, leading to substantial resource and time consumption. In contrast, feedback obtained from surrogate predictor\cite{nsganetv2} and lookup tables\cite{fbnet} based on offline perception is real-time and reusable. These predictors and tables can be constructed in advance and do not require fitting during the search process. Therefore, in this work, we fit a surrogate predictor by randomly sampling subnet structures to obtain a real-time and reusable resource proxy model. This accelerates the efficiency of neural architecture search.

\section{METHOD}
\begin{figure*}[t]
    \centering
    \includegraphics[width=1.0\textwidth]{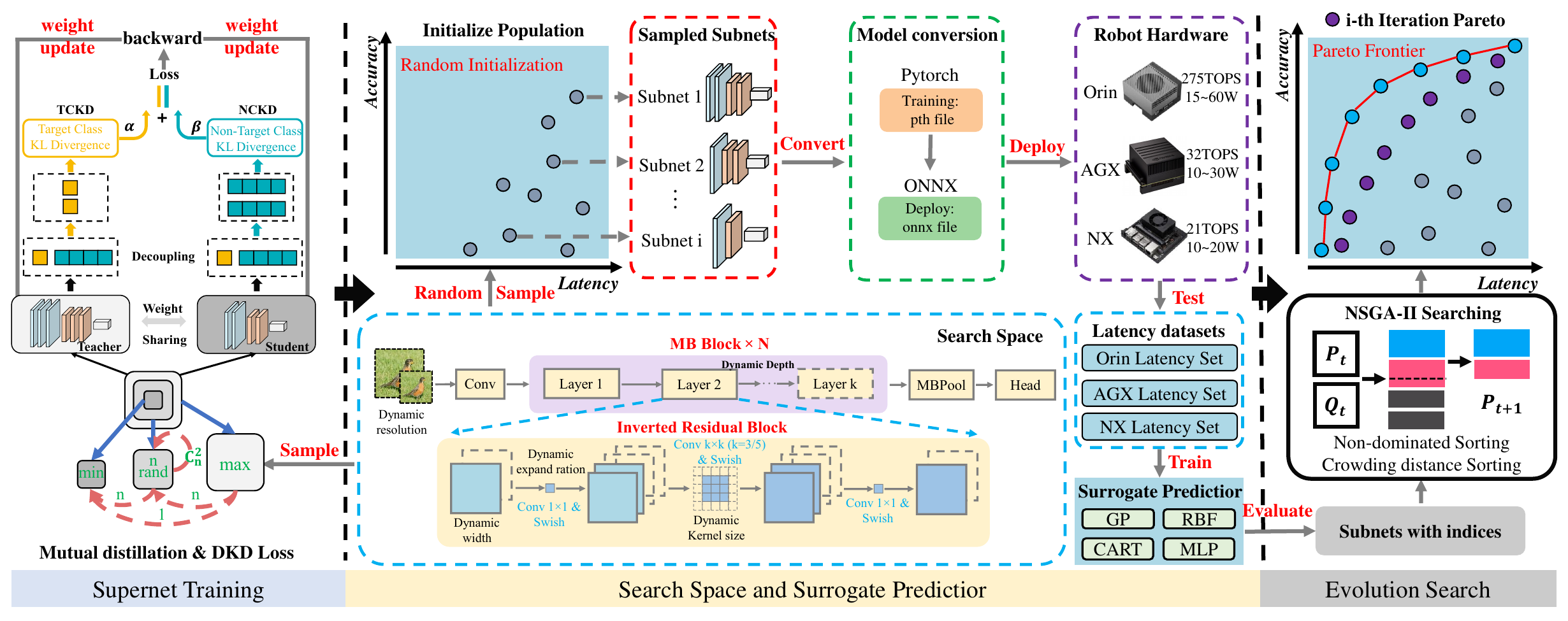}
    \caption{
    RAM-NAS employs the Sandwich Rule\cite{bignas} for subnet sampling, facilitating mutual distillation among diverse subnets and computing their DKD Loss. The supernet is then trained via backpropagation and weight updates. Following this, subnets are randomly sampled from the supernet and deployed on robot hardware for testing. This process generates Surrogate Predictor datasets containing inference latency, which are subsequently utilized to train the Surrogate Predictor.
    Finally, utilizing the NSGA-II multiobjective optimization algorithm, we define specific visual task accuracy and latency on target hardware as optimization objectives for the searched subnet. Through iterative processes, the Pareto frontier is derived, producing a set of Pareto-optimal solutions. 
    This set includes multiple optimal solutions simultaneously, enabling the generation of precision-optimal models that cater to varying computational latency requirements.}
    \vspace{-2em}
    \label{fig:pipline}
\end{figure*}

RAM-NAS consists of four main components: Preliminary, Improving the Supernet Training Strategy, Latency Surrogate Predictor, and Multiobjective Evolution Search under Resource-aware. Fig \ref{fig:pipline} provides a comprehensive overview of our entire approach.

\subsection{Preliminary}
{\bf{Problem Formulation.}} The weight-sharing NAS can be represented as a bi-level optimization problem in the following manner:

\begin{equation}
    \begin{aligned}
    \vspace{-4pt}
            & \alpha^* = \arg \max_{\alpha \in \mathcal{A}} \text{Acc}_{\text{val}}(\alpha, \mathbf{w}_\alpha^*), \\
             & \mathbf{w}_\alpha^* = \mathop{\arg\min}\limits_{\mathbf{w} \in \mathbf{W}} L_{\text{train}} (\alpha,\mathbf{w}_{\alpha}), 
    \end{aligned}
    \label{eq:NAS_preliminary}
\end{equation}
$\mathcal{A}$ and $\mathbf{W}$ denote the search space and weight of the supernet. Similarly, 
 $\alpha$ and $\mathbf{w}$ denote the architecture and weight of subnets sampled from the supernet.
$\alpha^*$ denotes the optimal architecture in search space $\mathcal{A}$ and $\mathbf{w}_\alpha^*$ denotes optimal weight under the candidate architecture $\alpha$. $L_{\text{train}} (\cdot)$ and $\text{Acc}_{\text{val}}(\cdot)$ represent optimizing the weights of supernet while minimizing supernet loss on the training set and maximizing the accuracy on the validation set. 

The aforementioned NAS method solely considers accuracy on the validation set, neglecting the need to account for inference delay during model deployment. Consequently, we formulate the NAS method as a multi-objective optimization problem. The formulation for this process is as follows:
\begin{equation}
    \begin{aligned}
        & F(\alpha)=(f_1(\alpha, \mathbf{w}_\alpha^*),\ldots,f_m(\alpha, \mathbf{w}_\alpha^*)), 
    \end{aligned}
\end{equation}
$F(\cdot)$ encompasses $m$ competing objectives, $f_i(\cdot)$ denotes the objective function, such as accuracy performance, FLOPs, and latency on specific hardware, etc. In this paper, we focus on two objectives: performance on validation set and latency on robot hardware.

{\bf{Search Space.}} Consistent with the previous One-Shot NAS method\cite{slimmable}, we need to transform the search space into a supernet, that covers all subnets with diverse configurations. 
To broaden the scope of the search and explore a wider range of possible solutions, we have chosen to adopt a broader search space built upon MobileNetV3, building upon the previous work\cite{attentivenas}.
MobileNetV3 is comprised of stacked blocks incorporating Inverted residual blocks, encompassing the five crucial dimensions of the convolutional neural network, namely resolution, width, expansion ratio, depth, and kernel size. Additionally, we have incorporated squeeze-and-excitation modules into select blocks. MBConv refers to an inverted residual block. SE denotes the squeeze and excites layer. width represents the channels per layer. Depth is the number of repeated MBConv blocks. Expansion ratio and kernel size denote the expansion ratio for depth-wise convolution layer and filter size. A comprehensive overview of the search space can be found in Table \ref{table:1}.

\begin{table*}[!]
\vspace{0.5em}
\begin{center}
\caption[l]{
{MobileNetV3-based search space. }}
\vspace{-0.4em}
\label{table:1}
\renewcommand{\arraystretch}{1.3}
\setlength{\tabcolsep}{2.9mm}{
\begin{tabular}{cccccccc}
\hline
\textbf{Stage} & \textbf{Block} & \textbf{Width} & \textbf{Depth} & \textbf{Kernel size}  & \textbf{Stride}& \textbf{Expand ratio} & \textbf{SE}  \\\hline
First Conv     & Conv           & \{16, 24\}     & -              & 3           & 2         & -                     & -            \\
Stage1         & MBConv         & \{16, 24\}     & \{1, 2\}       & \{3, 5\}     & 1               & 1                     & N            \\
Stage2         & MBConv         & \{24, 32\}     & \{3, 4, 5\}       & \{3, 5\}   & 2               & \{4, 5, 6\}                     & N            \\
Stage3         & MBConv         & \{32, 40\}     & \{3, 4, 5, 6\}     & \{3, 5\}   & 2                 & \{4, 5, 6\}                       & Y            \\
Stage4         & MBConv         & \{64, 72\}     & \{3, 4, 5, 6\}      & \{3, 5\}   & 2                 & \{4, 5, 6\}                       & N            \\
Stage5         & MBConv         & \{112, 120, 128\}     & \{3, 4, 5, 6, 7, 8\}       & \{3, 5\}    & 1                & \{4, 5, 6\}                       & Y            \\
Stage6         & MBConv         & \{192, 200, 208, 216\}     & \{3, 4, 5, 6, 7, 8\}       & \{3, 5\}     & 2              & 6                     & Y            \\
Stage7         & MBConv         & \{216, 224\}     & \{1,2\}       & \{3, 5\}        & 1             & 6                     & Y            \\
MBPool      & Conv           & \{1792, 1984\}      & -              & 1          & 1         & 6                     & -            \\\hline
\multicolumn{2}{c}{Input Resolution} & \multicolumn{6}{c}{\{192, 224, 276, 288\}} \\\hline
\vspace{-4em}
\end{tabular}}
\end{center}
\end{table*}

\subsection{Improving the Supernet Training Strategy}
Our supernet is constructed from the maximum value in the search space. Following the previous One-Shot method\cite{spos}, the subnets need to inherit the weight of the supernet to evaluate its performance to guide the optimal network selection during the search process. Therefore, comprehensive and high-quality training in supernet is very important for NAS. We solve this problem from two aspects: Subnet Mutual Distillation and Decoupled Knowledge Distillation.

\begin{figure}[!]
    \centering
    \includegraphics[width=0.49 \textwidth]{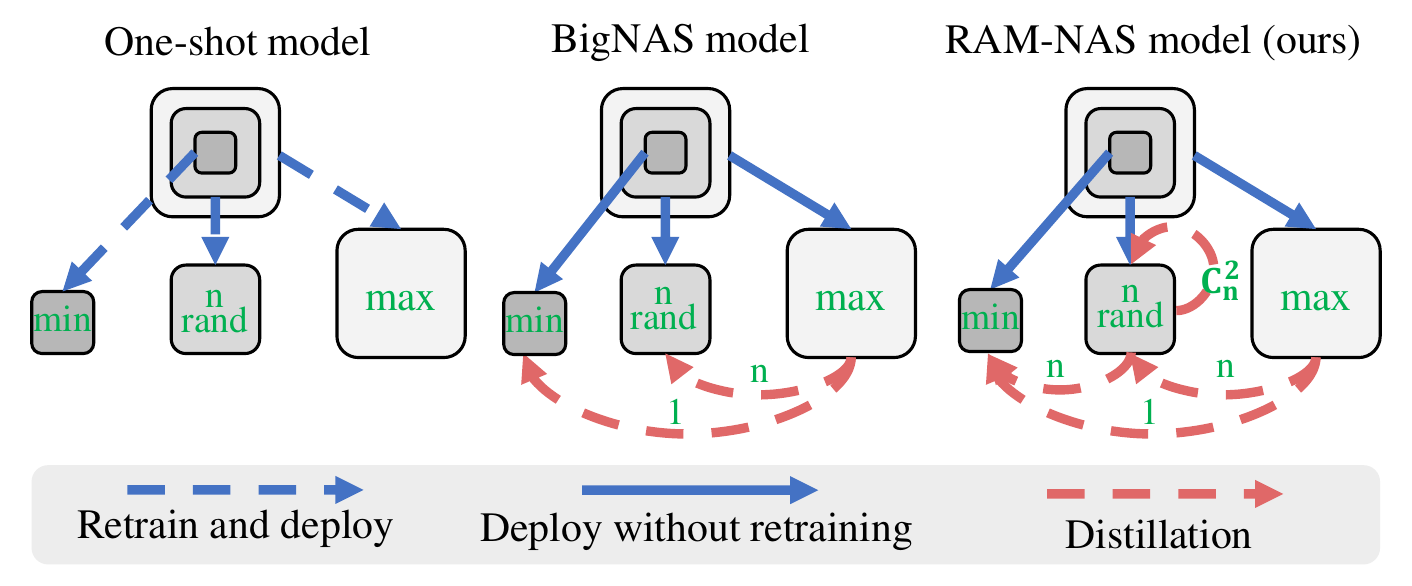}
    \caption{Comparison with other existing workflows. We employ nested squares to represent models with shared weights, use squares of different sizes to denote subnets at various scales, and use max, N rand, and min to indicate sampled subnets. The numbers within red dotted lines signify the count of distillations for connected subnets.}
    \vspace{-2em}
    \label{fig:mutual_distillation}
\end{figure}

{\bf{Subnets Mutual Distillation.}} Retrain-free NAS method\cite{bignas} proposes a new solution for training multiple networks simultaneously,  it can directly slice subnets of different sizes from the supernet and train them synchronously during the training process of the supernet. However, training multi-networks synchronously may cause gradient conflict of different subnets, resulting in the accuracy gap between single network and multi-networks\cite{pcgrad}. Therefore, BigNAS uses in-place distillation and sandwich rules to achieve supernet training. However, there is an issue with the in-place distillation. It uses the max subnet to distill two random subnets and the min subnet. Theoretically, it cannot distillate the knowledge between random sampling networks and the minimal network. A naive solution is all subnets distill each other, which can provide different perspectives and knowledge compared to relying on the max subnet to distill others. To address this problem, we mutually distill all the networks extracted by the sandwich rule in the pre-training stage of the supernet. The workflow is illustrated in Fig \ref{fig:mutual_distillation}.

In the supernet training phase, we apply the sandwich rule to sample the minimum model, maximum model, and N random models between the minimum and maximum (with N set to 2 in our experiments, following the approach in BigNAS\cite{bignas}). Each sampled subnet serves as a teacher for knowledge distillation to other subnets, enabling the calculation of their respective logits losses. The formulation for this process is as follows:
\begin{equation}
    \vspace{-4pt}
    \begin{split}
            &L_{\text{sum}}=L_{\text{max}}+\sum_{i=1}^{4}L_{\text{sub}_{i}}, \\
            &L_{\text{sub}_i} = \frac{1}{k-i}\sum_{j=i+1}^{k}D_{\text{KL}}(p_i||p_j),
            \label{eq:loss_compute}
    \end{split}
\end{equation}
where $L_{\text{max}}$  denotes the loss of ground truth with maximum subnet and $L_{\text{sub}_{i}}$ denotes the $i\text{-}th$ subnet with other subnets. $D_{\text{KL}}(p_i||p_j)$ denotes the KL divergence between the prediction logits $p_i$ and $p_j$ of teachers and students, it takes into account the remaining $k-i$ subnets as imitated teachers, calculates their KL divergence, and then computes the average. 

Due to the weight-sharing strategy, multiple forward-backward propagation processes are necessary for each randomly sampled subnet to distill knowledge from one another. Every sampled subnets $\alpha_i \subseteq \mathcal{A}$ and its weight $\mathbf{w}_i \subseteq \mathbf{W}$.
It subsequently aggregates all gradients from the ground truth with the maximum subnet and all mutual distillers before updating the weights. The gradient aggregation process can be formulated as follows:
\begin{equation}
    \begin{split}
        &  \mathbf{W} = \begin{cases} 
            \mathbf{W} + \gamma \displaystyle\frac{\partial L_{\text{max}}}{\partial \mathbf{w}}, & \text{all weights} \\
            \mathbf{W} + \gamma \displaystyle\frac{\partial L_{\text{sub}_{i}}}{\partial \mathbf{w}}, & \text{if \;} \mathbf{w} \in \mathbf{W}_{stu} \\
            \mathbf{W}, & \text{if \;} \mathbf{w} \notin \mathbf{W}_{stu} 
        \end{cases} 
        \label{eq:weight_update}
    \end{split}
\end{equation}
where $\gamma$ denotes the learning rate and $\mathbf{W}_{stu}$ denotes the weight belonging to the student model in distillation. During the gradient update phase, we solely update the gradient associated with the student model and keep the weights of other components fixed. Since both the randomly sampled subnets and min-max subnets are included in the gradient aggregates, mutual distillation can update the weights of all dimensions throughout the iteration.

\begin{figure}[!]
\vspace{0.5em}
    \centering
    \includegraphics[width=0.49\textwidth]{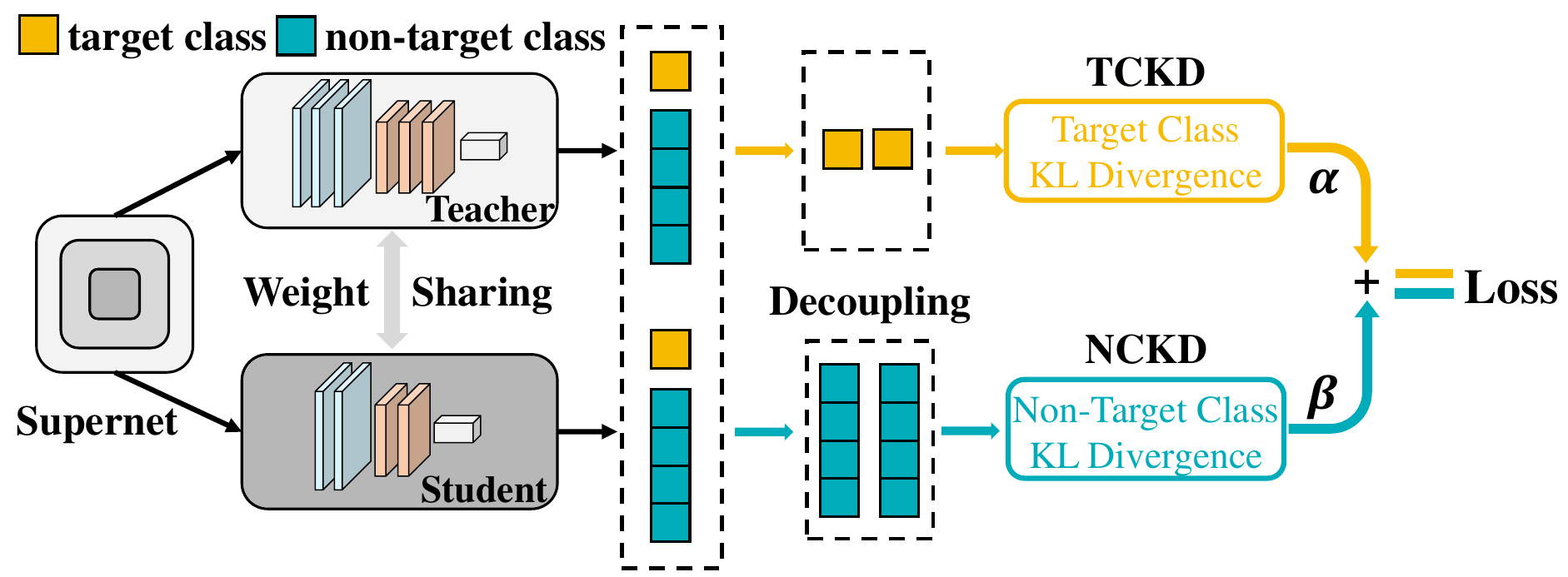}
    \caption{Illustration of the DKD. It couples the importance of two parts about knowledge distillation: the loss of both the target class and the non-target class.}
    \vspace{-2em}
    \label{fig:DKDLoss}
    
\end{figure}

{\bf{Decoupled Knowledge Distillation.}} Previous research\cite{fitnets} has demonstrated that feature distillation consistently outperforms logits-based distillation across various tasks. However, the varying depths of subnets within our supernet pose a challenge in determining the state of the intermediate feature layer during mutual distillation. This makes it challenging to extract features from the feature extraction layers.
To enhance the effectiveness of logits distillation, a method proposed in \cite{dkd} is adopted. This method involves splitting the traditional KD loss into two components: target class knowledge distillation and non-target class knowledge distillation. Each component is then calculated separately and their losses are combined using a weighted sum. The workflow for this approach is depicted in Fig \ref{fig:DKDLoss}.
The DKD method enables the attainment of greater flexibility and efficiency in knowledge distillation, while also addressing the issue of the non-target class being overshadowed by the target class in traditional distillation methods. The formula can be represented as follows:
\begin{equation}
    \begin{split}
            &L_{\text{DKD}} = \alpha \times L_{\text{TC}}+\beta \times L_{\text{NC}}, \\ 
            &L_{\text{TC}} = D_{\text{KL}}(\boldsymbol{c}^T\|\boldsymbol{c}^S), \\ 
            &L_{\text{NC}} = D_{\text{KL}}(\hat{\boldsymbol{c}}^{T}\|\hat{\boldsymbol{c}}^{S}),
            \label{eq:DKDLoss}
    \end{split}
\end{equation}
where $\alpha$ and $\beta$ denotes the weight of target class loss $L_{\text{TC}}$ and non-target class loss $L_{\text{NC}}$, respectively. 
$\boldsymbol{c}^T$ and $\boldsymbol{c}^S$ denote the probability distributions of the target class for the teacher and the student, calculated using the softmax function. Similarly, $\hat{\boldsymbol{c}}^T$ and $\hat{\boldsymbol{c}}^S$ represent the probability distributions of the non-target classes for the teacher and the student, also calculated using the softmax function.

\subsection{Latency Surrogate Predictor}
After the supernet training is completed, a multiobjective evolutionary search is needed to obtain the Pareto frontier. To reduce the computational burden caused by evaluating subnets performance, we use surrogate predictors to assist neural architecture search and speed up the fitness evaluation process. 
We first random sample subnets $\alpha_{i}$ from pre-trained supernet search space $A$ and get its encoding format, then using corresponding weights to evaluate the accuracy $f_i$ on the validation set. 
To ensure real-time execution of the searched subnets on the robot hardware platform, our objective is to minimize the inference delay on the hardware platform. Nevertheless, the intricate calculations involved in the search process necessitate its execution on the server platform.
As a consequence, it is unable to consider the inference latency of the actual robot hardware. To address this limitation, we employ a latency surrogate predictor to predict the actual inference latency on the target hardware. For the reusability of the latency surrogate predictor, we randomly sample the subnet $\alpha_{i}$ and obtain the inference latency indicator $l_{i}^{n}$ on the target platforms, where $n$ is the $n$-th hardware platform,We randomly sample subnets $\alpha_{i}$ and obtain the inference latency metric $l_{i}^{n}$ on the target platform. The implementation of this part is shown in section IV, part A.
Latency datasets $(\alpha_{i}, l_{i}^{n})$ are then constructed for training the latency surrogate predictor. Subsequently, the latency surrogate predictor is trained.
This predictor can establish a mapping relation between latency and encoding format, and the predictor drives the evolutionary search. 
However, as indicated in \cite{tai}, the efficacy of the surrogate predictor heavily relies on the accuracy of its predictions. Therefore, the fulfillment of the following two properties is crucial:
(1) High-rank correlation between predicted performance and true performance.
(2) Sample efficiency: Since obtaining each sample requires expensive subnet inference evaluation on hardware, it is necessary to minimize the number of samples required to train the surrogate predictor.


\subsection{Multiobjective Evolutionary Search under Resource-aware}
We employ the NSGA-II algorithm\cite{nsga} as the standard multiobjective algorithm to explore the trade-off between model accuracy and hardware latency, two conflicting objectives. The pseudo-code for this method is provided in Algorithm \ref{alg1}. First, the initial population size $P$ is obtained through random initialization, and then the dominance relationship of each individual in the population is calculated and the non-dominated solution is determined. Based on the non-dominance, the solution is divided into different frontier levels. At the same time, in order to prevent the solution distribution density from being too large and maintain the diversity of the Pareto frontier, sorting is performed by calculating the crowding degree, and finally, the solution is obtained based on the non-dominated sorting and crowding degree calculation results. NSGA-II uses a standard genetic algorithm, at the beginning of the genetic algorithm, $P$ architectures are randomly selected as seeds. The top $K$ architectures are selected as the parent generation, and the next generation is generated through crossover and mutation. For crossover, two randomly selected candidates and crossed to produce a new candidate in each generation. For mutation, the candidate mutates with probability $P_m$ to produce a new architecture. The next generation is generated through crossover and mutation, which helps to retain the elite individuals in the Pareto frontier. As the algorithm iterates, a Pareto frontier is eventually formed and non-dominated solutions are output from the pool of evaluated architectures.

\begin{algorithm}
\caption{Multiobjective Evolutionary Search}
\label{alg1}
\begin{algorithmic}[1]
\renewcommand{\algorithmicrequire}{\textbf{Input:}}
\renewcommand{\algorithmicensure}{\textbf{Output:}}
\REQUIRE 
{search sapce $A$, supernet weights $W_{A}$, population size $P$, number of generation iteration $T$, mutation probability $p_m$, top $K$ architectures.}
\ENSURE{The  searched architecture $\alpha^*$ with $w^{*}$ on the Pareto frontier.}
\STATE Random sample $P_1$ subnet architectures \{$\alpha_1, \alpha_2, \ldots, \alpha_P$\} from $A$ and its weights $\{w_1, w_2, \ldots, w_p\} \in W_{A}$
\STATE Latency predictor $pred_{lat} \leftarrow fit(\alpha_i, l_i^{n})$
\STATE $n := P/2$, $m := P/2$
\WHILE{search step $t \in (2,T)$} 
    \STATE $\text{Acc} \leftarrow eval(\alpha_i,w_i), \text{Lat} \leftarrow pred_{lat}(\alpha_i,w_i)$ 
    \STATE $P_{t} \leftarrow \text{NSGA-II}(\text{Acc},\text{Lat}, \alpha_i,w_i)$
    \STATE $K \leftarrow \text{Update-Topk}(P_{t})$
    \STATE $P_{t+1}^{c} \leftarrow \text{Crossover}(K, n) $
    \STATE $P_{t+1}^{m} \leftarrow \text{Mutation}(K, m, p_m) $ 
    \STATE $P_{t+1} \leftarrow P_{t+1}^{c} \cup P_{t+1}^{m}$
\ENDWHILE
\RETURN Pareto frontier architecture $\alpha^*$ with $w^{*}$
\end{algorithmic}
\end{algorithm}

\section{EXPERIMENT}
In this section, we evaluate the performance of RAM-NAS on various robot vision tasks. In addition, we conducted a comparison of results achieved through FLOPs and resource-aware search, assessing the inference latency on the robotic hardware platform. 
Finally, we conduct ablation experiments to verify our improved supernet training strategy and evaluate the correlation and sampling efficiency of surrogate predictors for latency.

\subsection{Implementation Details}
{\bf{Supernet Training and Downstream Tasks Finetune.}}
The details of these training runs can be found in Table \ref{supernet_train}. We conducted all training sessions on 8 GPUs (NVIDIA GeForce RTX 3090) using PyTorch 1.11. For image classification, a mini-batch size of 448 was employed, whereas for object detection, it was 144, and for semantic segmentation, it was 8. 
To better adapt our method to robot operational scenarios, we applied series data augmentation strategies, such as brightness adjustment, noise addition, and random mask.
Unless explicitly stated otherwise, all other configurations were kept consistent with those outlined in BigNAS\cite{bignas}.

\begin{table}[t]
	\caption{Supernet pre-train and downstream tasks finetune settings. \label{supernet_train}
	}
 \renewcommand\arraystretch{1.1}
	\centering
	\resizebox{0.47\textwidth}{!}{
		\begin{tabular}{cccccccc}
			\toprule
			Tasks &Epochs & \begin{tabular}[c]{@{}c@{}} Warmup \\ Epochs \end{tabular} & Optimizer & Lr   & \begin{tabular}[c]{@{}c@{}}Lr \\ scheduler\end{tabular} & \begin{tabular}[c]{@{}c@{}}Weight\\ Decay\end{tabular} & Dataset \\ \midrule
			Classification &240 &10 & SGD     & 0.8 & cosine & 1e-4 & ImageNet-1K \\ 
            Detection &20 &1 & SGD     & 0.2 & cosine & 1e-4 & COCO\\ 
            Segmentation &20 &1 & SGD     & 0.2 & cosine & 1e-4 & Cityscapes\\ 
   \bottomrule
	\end{tabular}}
 \vspace{-14pt}
\end{table}

{\bf{Robotic Hardware Resource-aware.}} In our latency surrogate predictor fitting and resource-aware experiments, we use three Nvidia-produced robotic hardware types: Nvidia Jetson AGX Orin, Xavier, and Nvidia Xavier NX. We use ONNX, an open format designed for model representation, to measure the latency of these models on three different types of hardware.  It's important to mention that all models utilize onnxruntime-gpu for inference, with the ONNX component CUDAExecutionProvider and fp32 employed to accelerate the inference process.

{\bf{Latency Surrogate Predictor Fitting.}} Following\cite{nsganetv2}, we employ four distinct surrogate predictors for prediction: Multi-Layer Perceptron (MLP), Classification And Regression Trees (CART), Radial Basis Function (RBF), and Gaussian Process (GP). First, we sampled 3000 subnets from the supernet, and their latencies were measured across three types of hardware. The dataset was split, with $80\%$ used for training and $20\%$ reserved for testing purposes.
Subsequently, we choose the most effective predictor on the test dataset as our surrogate predictor.

{\bf{Multiobjective Evolution Search.}} To initiate the process of evolutionary search, we select a population size of $5120$ random architectures, denoted as $P$. Subsequently, the top $2560$ architectures from this population are chosen as parents to generate the next generation through crossover and mutation. Additionally, to promote diversity within the population, a candidate may undergo mutation with a probability of $P_m$ to generate a new architecture. In all of our experiments, we have set $P_m$ to $0.2$. The mutation and crossover operations are iteratively performed until the population size $p$ is reached with a sufficient number of new candidates. The maximum number of iterations $T$ is set to $30$.

\subsection{Classification task}
Classification is a basic task to prove the effectiveness of our supernet training strategy. To facilitate comparison with previous methods, we first set the optimization goal to top-1 accuracy and FLOPs on the validation data, demonstrating the benefits of our supernet pre-training strategy. The Supernet encompasses sizes ranging from $200$ to $2000$ MFLOPs, within which we explore architectures spanning from $200$ to $1000$ MFLOPs. Table \ref{tab:compare_with_sota} shows the results on ImageNet, establishing new accuracy vs. FLOPs trade-offs. Under constraints ranging from $200$-$1000$ MFLOPs, our method surpassed all competitors. For small-sized models, our RAM-NAS-A0 achieves $76.7\%$ accuracy under only $211$ MFLOPs, which is $1.5\%$ higher than MobileNetV3 $1.0\times$ with fewer FLOPs. In the case of EfficientNet, our RAM-NAS-A3 achieves a Top-1 Accuracy of $79.1\%$ with a significantly lower requirement of $460$ MFLOPs, as compared to the $700$ MFLOPs needed by EfficientNet-B1. In contrast to previous NAS methods, our approach yields superior outcomes. For instance, when compared to BigNAS-S, our method attains a Top-1 accuracy of $76.7\%$ using only $211$ MFLOPs, whereas BigNAS-S necessitates $242$ MFLOPs to achieve a Top-1 accuracy of $76.5\%$.

\begin{figure}[!]
    \centering
    \includegraphics[width=0.49\textwidth]{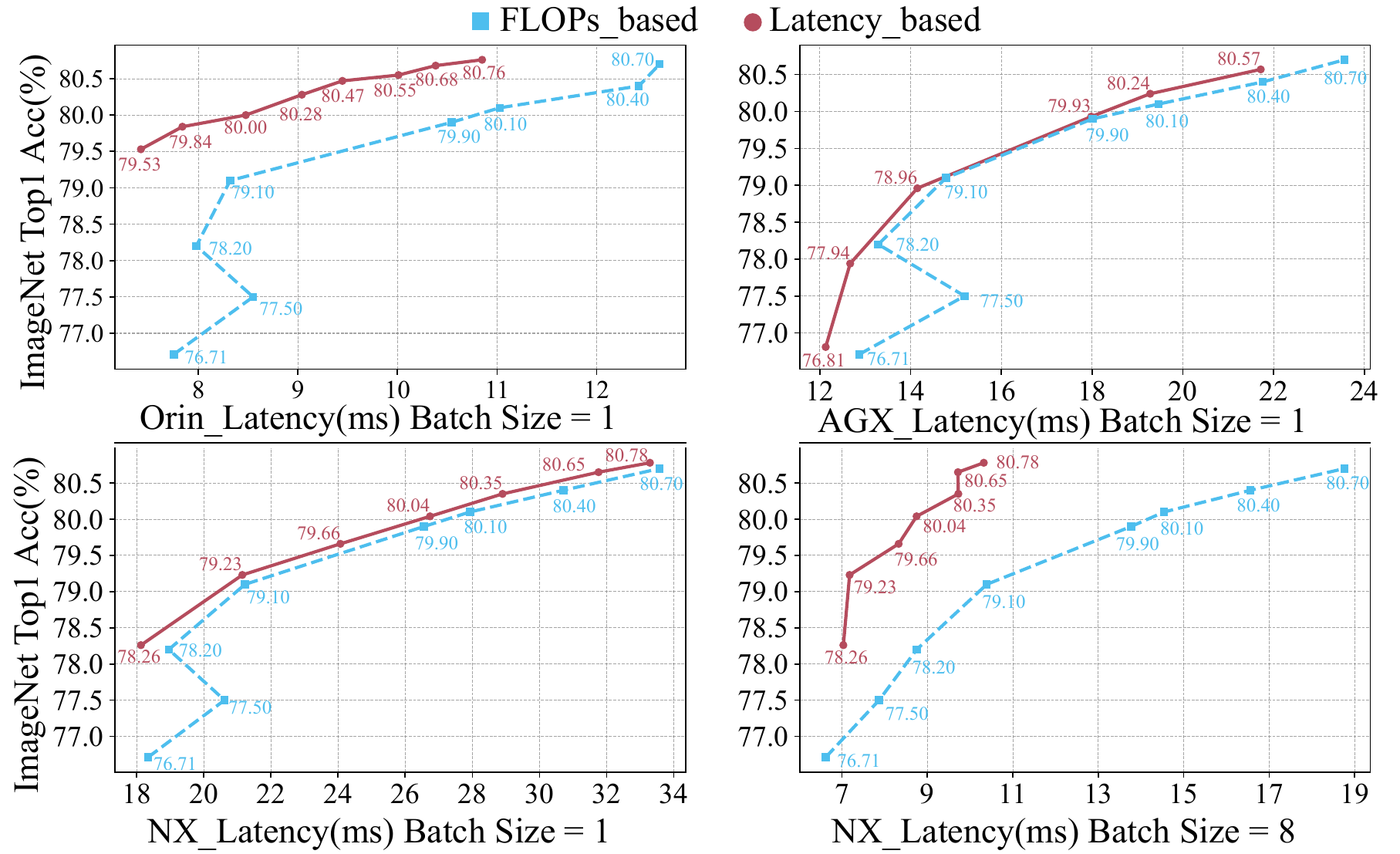}
    \caption{Illustration of RAM-NAS with resource-aware objectives and FLOPs objectives on three robotic hardware.}
    \vspace{-2em}
    \label{fig:latency_vs_flops}
\end{figure}

Besides, we run RAM-NAS with resource-aware objectives: maximize Top-1 accuracy on ImageNet validation data and minimize latency on the hardware platforms.
As illustrated in Fig \ref{fig:latency_vs_flops}, the FLOPs-based models correspond to RAM-NAS-A0 through RAM-NAS-A7. The Latency-based models are discovered using the latency surrogate predictor that is capable of recognizing hardware resources. To better align with the operational requirements in robotics, we have standardized the batch size for inference to $1$. Compared to models that rely on the search for FLOPs as optimization targets, our resource-aware method can search for models with smaller inference latency and higher accuracy on each target hardware. 
Simultaneously, experimental results indicate a lack of precise mapping between the FLOPs of a model and the actual hardware inference delay. Surprisingly, models with larger FLOPs exhibit lower inference delays, challenging the conventional understanding. For example, in Fig \ref{fig:latency_vs_flops}, RAM-NAS-A1 has fewer FLOPs than RAM-NAS-A2, but its inference latency is higher across all hardware platforms when the batch size is 1.
Meanwhile, the disparity is more pronounced on Orin, whereas the margin is narrower on AGX and NX, since the DataLoader may be time-consuming. Hence, we experimented using a batch size of 8 on the NX, which possesses the lowest computing power. Our findings indicate that, with an increase in batch size, the Latency-based search model exhibits significant improvement compared to the FLOPs-based model.

\begin{table}[ht]
    \centering
    \setlength{\tabcolsep}{1pt}
    \caption{Comparison with prior manual design and NAS approaches on ImageNet.}
    \begin{tabular}{clcc}
        \hline 
        Group & Method & FLOPs(M) & Top-1$(\%)$  \\
        \hline 
    \multirow{13}{6em}{200-400 (M) } 
    & \textbf{RAM-NAS-A0}     &\textbf{211}     &\textbf{76.7}            \\ 
        & MobileNetV3 {\scriptsize 1.0$\times$}~\cite{mobilenetv3} & 217 & 75.2 \\
        & OFA~\cite{OFA} & 230 & 76.0 \\
        & FBNetV2-F4~\cite{fbnetv2} & 238 & 76.0 \\
        & BigNAS-S~\cite{bignas} & 242 & 76.5 \\
    & \textbf{RAM-NAS-A1}     &\textbf{284}     &\textbf{77.5}            \\ 
        & MNasNet-A1~\cite{mnasnet} & 312 & 75.2 \\
        & ProxylessNAS~\cite{proxylessnas} & 320 & 74.6 \\
        & FBNetV2-L1~\cite{fbnetv2} & 325 & 77.2 \\
    & \textbf{RAM-NAS-A2}    &\textbf{325}     &\textbf{78.2}             \\
        & FBNetV3~\cite{fbnetv3} & 343 & 78.0 \\
        & MobileNetV3 {\scriptsize 1.25$\times$}~\cite{mobilenetv3} & 356 & 76.6 \\
        & EfficientNet-B0~\cite{efficientnet} & 390 & 77.1 \\  \hline
    \multirow{6}{6em}{400-600 (M) }
        & MNasNet-A3~\cite{mnasnet} & 403 & 76.7 \\
        & BigNAS-M~\cite{bignas} & 418  &  78.9 \\
        & FBNetV2-L2~\cite{fbnetv2} &  422 & 78.1\\ 
        & \textbf{RAM-NAS-A3}     &\textbf{460}   &\textbf{79.1}  \\ 
        & BigNAS-L~\cite{bignas} & 586 & 79.5 \\  \hline 
    \multirow{4}{6em}{600-800 (M) }
    & \textbf{RAM-NAS-A4}     &\textbf{625}   &\textbf{79.9} \\
        & EfficientNet-B1~\cite{efficientnet} & 700 & 79.1 \\
    & \textbf{RAM-NAS-A5}     &\textbf{704}   &\textbf{80.1} \\\hline 

    \multirow{4}{5em}{$>$800 (M) } 
    & \textbf{RAM-NAS-A6}     &\textbf{808}   &\textbf{80.4} \\
    & \textbf{RAM-NAS-A7}     &\textbf{976}   &\textbf{80.7} \\
        & EfficientNet-B2~\cite{efficientnet} & 1000 & 80.1 \\ \hline 
    \end{tabular} 
    \label{tab:compare_with_sota}
    \vspace{-1.5em}
\end{table}

\subsection{Detection and Segmentation task}
Moreover, we apply the searched RAM-NAS models to object detection and semantic segmentation tasks to verify the scalability of our methods. For fair comparison, we replaced the original backbone with the model obtained through our search process and the weights were fine-tuned specifically for each task. For object detection, we followed the MobileNetV3\cite{mobilenetv3} architecture and used the RAM-NAS search-derived backbone directly as the feature extractor for SSDLite. In our experiments, the initial layer of SSDLite was connected to stage 4 in search space, and the second layer of SSDLite was connected to the top of the stage 7.  For semantic segmentation, the layer of LRASPP was connected to stage1, 2 and 7.

As shown in Table \ref{tab:downstream_det}, we present the mAP/mIOU scores for each method on their respective datasets and provide the inference latency on Orin. The A4 and A7 represent the flops searched model, while $79.84, 80.00, 80.76$ correspond to structures identified through the original latency surrogate model.
In particular, RAM-NAS-SSD (80.76) achieves 26.2 mAP, providing +$1.7$ mAP gain over RAM-NAS-SSD (A7), and increases latency by $0.5$ ms, 
RAM-NAS-LRASPP (80.00) achieves 68.70 mIOU, providing $+0.39$ mIOU gain over RAM-NAS-LRASPP (A7) and $1.27\times$ speedup on Orin. 

\begin{table}[]
\vspace{0.5em}
\centering
\caption{mean average precision(map)/mean intersection over union(miou) and latency on COCO and cityscapes dataset.}
\label{tab:downstream_det}
\setlength{\tabcolsep}{1.8mm}{
\begin{tabular}{cccccc}
\hline
Models  & mAP/mIOU(\%)  & Latency on Orin(ms) \\\hline
MobileNetV3-SSD    & 20.2 & 20.57  \\
RAM-NAS-SSD (A4)     & 23.6 & 24.73 \\
RAM-NAS-SSD (A7)     & 24.5 &26.98  \\
\rowcolor{green!20} RAM-NAS-SSD (80.00) & 23.3  & 23.43 \\
\rowcolor{green!20} RAM-NAS-SSD (80.76) & 26.2  & 27.48 \\

\hline
MobileNetV3-LRASPP    & 64.11 &53.53 \\
RAM-NAS-LRASPP (A4)     & 67.41 &95.30  \\
RAM-NAS-LRASPP (A7)     & 68.31 &124.42 \\
\rowcolor{green!20} RAM-NAS--LRASPP (79.84)     & 66.76 &82.95 \\
\rowcolor{green!20} RAM-NAS-LRASPP (80.00)     & 68.70 &98.43  \\
\hline
\vspace{-3em}
\end{tabular}}
\end{table}

\subsection{Ablation Study}
{\bf{Improving the Supernet Training Strategy.}} 
During the supernet pre-training phase, in order to figure out which component is working, we further performed the following ablation experiments. We evaluated their performance based on two dimensions: min subnet and max subnet accuracy on Imagnet, as they determine the lower and upper bounds of the supernet, as reported in Table \ref{tab:Ablation_1}. In Baseline, we employ in-place distillation and naive distillation loss\cite{kd} in BigNAS. In EXP1, we use subnets mutual distillation and naive distillation loss in our subnets. This approach results in a Top1 improvement of $0.4\%$ for the min subnet and $0.18\%$ for the max subnet. It aggregates knowledge and information across networks to maximize the lower bound of the network, enabling each network to acquire additional knowledge from other networks. In EXP2, we use naive distillation and Decoupled Knowledge Distillation, the $\alpha$ is set as $1.0$, and $\beta$ is set as $0.5$. It yields a $0.28\%$ boost and $0.17\%$ over baseline. Compared to using the default BigNAS settings\cite{bignas}, utilizing SMD and DKD separately notably enhances the performance of the min subnet and max subnet. Combining both further enhances the performance of the min subnet and max subnet. Our supernet training strategy surpasses the baseline $0.44\%$ and $0.36\%$ with the same structure.

\begin{table}[!h]
\centering
\renewcommand\arraystretch{1.2}
    \caption{\textbf{Ablation study results about the improved supernet Training Strategy on ImageNet.} SMD: Subnets Mutual Distillation. DKD: Decoupled Knowledge Distillation.}
    \resizebox{0.47\textwidth}{!}{
        \centering
        \begin{tabular}{l|cc|ll|ll}
            \toprule
            & \multicolumn{2}{c|}{Methods} & \multicolumn{2}{c|}{min subnet} & \multicolumn{2}{c}{max subnet} \\ 
            \cline{2-7}
             & SMD & DKD   & Top1$(\%)$   & Top5$(\%)$      & Top1$(\%)$   & Top5$(\%)$     \\ \midrule
            Baseline       &                 &                   & 76.24            & 92.76     & 81.03             & 95.10       \\
            EXP1       &$\checkmark$     &                   &76.64\textsubscript{\color{red} (+0.40)}               & \textbf{92.96}\textsubscript{\color{red} (+0.20)}     & 81.21\textsubscript{\color{red} (+0.18)}              & 95.21\textsubscript{\color{red} (+0.11)}     \\
            EXP2       &                 &$\checkmark$       & 76.52\textsubscript{\color{red} (+0.28)}              & 92.89\textsubscript{\color{red} (+0.13)}     & 81.20\textsubscript{\color{red} (+0.17)}             & 95.18\textsubscript{\color{red} (+0.08)}       \\
            EXP3       &$\checkmark$     &$\checkmark$       & \textbf{76.68}\textsubscript{\color{red} (+0.44)}    & 92.95\textsubscript{\color{red} (+0.19)}    & \textbf{81.39}\textsubscript{\color{red} (+0.36)}             & \textbf{95.26}\textsubscript{\color{red} (+0.16)}    \\ 
              \bottomrule
    \end{tabular}}
    \vspace{-4pt}
    \label{tab:Ablation_1}
\end{table}

\begin{figure}[!]
    \centering
    \includegraphics[width=0.49\textwidth]{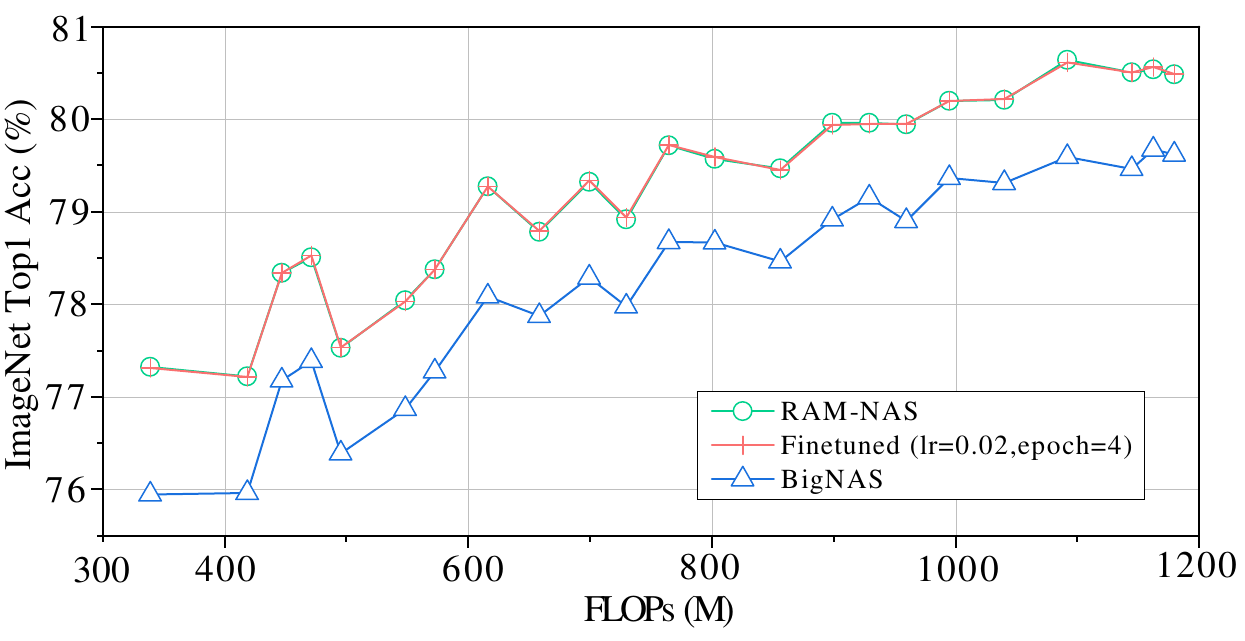}
    \caption{Illustration of the consistency between subnet directly searched by BigNAS\cite{bignas}, our method, and finetuned performance.}
    \vspace{-2.2em}
    \label{fig:consistency}
\end{figure}

Furthermore, to demonstrate the effectiveness of our proposed supernet training strategy in enhancing the consistency between the directly searched subnets and the performance after fine-tuning, we randomly sampled subnets with varying FLOPs. 
To attain optimal subnet performance, the RAM-NAS sampled subnets undergo fine-tuning with a learning rate of $0.02$ over 4 epochs to achieve their actual performance.
At the same time, the random seeds are fixed and the supernet trained by BigNAS is randomly sampled. The results are as shown in Fig \ref{fig:consistency}.
The figure illustrates that subnets obtained through our supernet training strategy exhibit comparable or even superior performance compared to fine-tuned results across different FLOPs, with an RMSE of just $0.067$ about Finetuned and RAM-NAS. 
Meanwhile, RAM-NAS outperforms BigNAS in subnets of varying FLOPs. This result demonstrates the effectiveness of our approach in benefiting subnets at all levels within the supernet. It enhances the generalization of subnets and helps discover more robust values for weight-sharing supernet parameters.

\begin{figure}[!]
    \centering
    \includegraphics[width=0.49\textwidth]{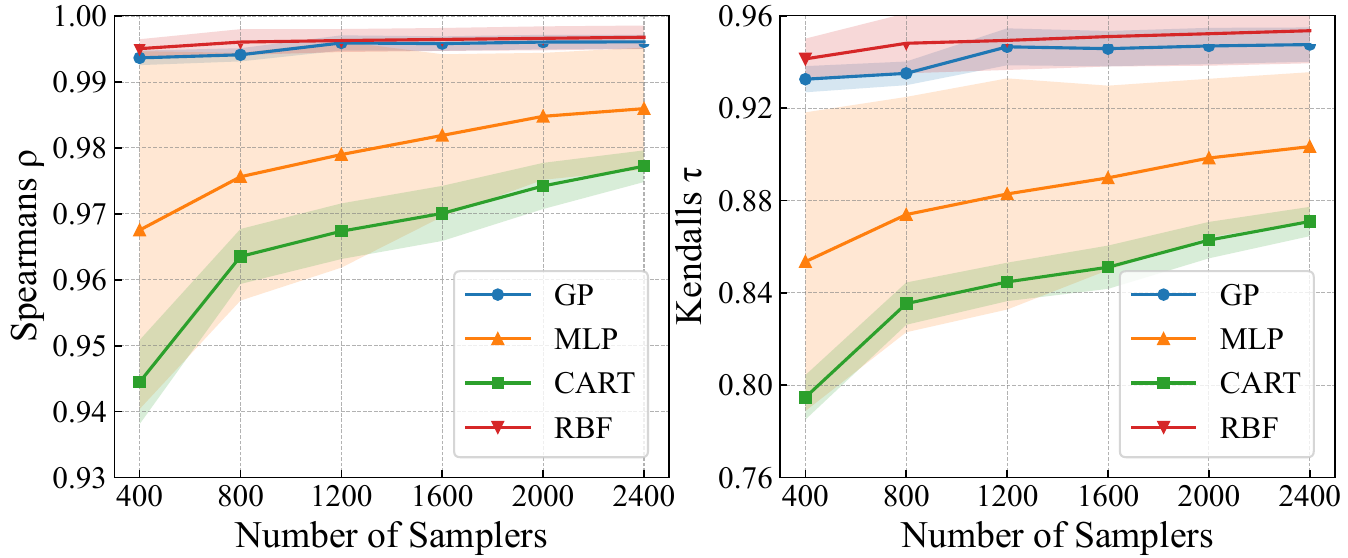}
    \caption{Illustration of the latency predictors ranking correlation and sampling efficiency.}
    \label{fig:lat}
    \vspace{-2.2em}
\end{figure}

{\bf{Performance of the Surrogate Predictors.}} 
We use Spearman correlation coefficient $\rho$ and Kendall correlation coefficient $\tau$ to quantitatively evaluate the ranking correlation of surrogate predictors. These indicators have a value range of [-1, 1]. A value closer to 1 indicates a higher prediction performance.
In addition to comparing correlation coefficients, sample efficiency is also used as an evaluation metric for Surrogate Predictors. We replicate the previous experiment and systematically decrease the number of samples used to train the Predictors.
Then, we evaluated the correlation coefficients and sample efficiency across three distinct hardware setups and generated a dot-line plot illustrating the variances, as depicted in Fig \ref{fig:lat}.
The findings indicate that GP and RBF outperform other methods in predicting latency consistency and sampling efficiency. Even with a small sampling size, they consistently maintain a high ranking. However, RBF  exhibits a superior mean compared to GP. 
Therefore, our method can achieve excellent performance on new hardware with only a small number of samples for both sampling and training. 

\section{CONCLUSIONS}

In this paper, we introduced RAM-NAS, a resource-aware multiobjective evolutionary NAS method that focuses on improving the supernet pre-train and robot resource-aware. We propose a subnet mutual distillation to substitute the in-place distillation and utilize DKD Loss to achieve more flexible and efficient knowledge distillation. 
We also develop a surrogate predictor to achieving high-rank correlation and sample efficiency. Our approach is the first to incorporate robotic hardware resources into neural architecture search, achieving a trade-off of accuracy and latency through multiobjective optimization. 
Based on the RAM-NAS method, there is considerable potential for further research: (1) Exploring more complex neural architectures. (2) Conducting latency testing with a greater number of sensor inputs and higher resolutions. (3) Adapting  to other types of  robotics hardware platforms, such as x86, ARM, and NPU.

\bibliographystyle{IEEEtran}
\bibliography{myref}





\end{document}